\documentclass{article} 
\usepackage{colm2024_conference}

\usepackage{microtype}
\usepackage{hyperref}
\usepackage{url}
\usepackage{booktabs}
\usepackage[pdftex]{graphicx}
\usepackage{arydshln}
\usepackage[whole]{bxcjkjatype}

\title{Building a Large Japanese Web Corpus\\for Large Language Models}

\author{Naoaki Okazaki\footnotemark[2], Kakeru Hattori\footnotemark[2], Hirai Shota\footnotemark[2], Hiroki Iida\footnotemark[2], Masanari Ohi\footnotemark[2], \\ \textbf{Kazuki Fujii\footnotemark[2], Taishi Nakamura\footnotemark[2], Mengsay Loem\footnotemark[2], Rio Yokota\footnotemark[3], Sakae Mizuki\footnotemark[2]} \\
\footnotemark[2]\; Department of Computer Science, School of Computing, Tokyo Institute of Technology\\
\footnotemark[3]\; Global Scientific Information and Computing Center, Tokyo Institute of Technology\\
\; 2-12-1 Ookayama, Meguro-ku, Tokyo, 152-8550 Japan\\
\; \texttt{\{okazaki@c, kakeru.hattori@nlp.c, shota.hirai@nlp.c, hiroki.iida@nlp.c,} \\
\; \ \, \texttt{masanari.ohi@nlp.c, kazuki.fujii@rio.gsic, taishi.nakamura@rio.gsic,} \\
\; \ \, \texttt{mengsay.loem@nlp.c, rioyokota@rio.gsic, sakae.mizuki@nlp.c\}.titech.ac.jp}
}

\colmfinalcopy 
\begin{document}

\maketitle

\begin{abstract}
Open Japanese large language models (LLMs) have been trained on the Japanese portions of corpora such as CC-100, mC4, and OSCAR. However, these corpora were not created for the quality of Japanese texts.
This study builds a large Japanese web corpus by extracting and refining text from the Common Crawl archive (21 snapshots of approximately 63.4 billion pages crawled between 2020 and 2023).
This corpus consists of approximately 312.1 billion characters (approximately 173 million pages), which is the largest of all available training corpora for Japanese LLMs, surpassing CC-100 (approximately 25.8 billion characters), mC4 (approximately 239.7 billion characters) and OSCAR 23.10 (approximately 74 billion characters).
To confirm the quality of the corpus, we performed continual pre-training on Llama 2 7B, 13B, 70B, Mistral 7B v0.1, and Mixtral 8x7B Instruct as base LLMs and gained consistent (6.6--8.1 points) improvements on Japanese benchmark datasets.
We also demonstrate that the improvement on Llama 2 13B brought from the presented corpus was the largest among those from other existing corpora.
\end{abstract}

\section{Introduction}

ChatGPT, released by OpenAI in late 2022, established a milestone toward achieving general-purpose artificial intelligence.
Research topics using large language models (LLMs) attract much attention including chain of thoughts~\citep{NEURIPS2022_9d560961,NEURIPS2022_8bb0d291,wang2023selfconsistency,trivedi-etal-2023-interleaving}, instruction tuning~\citep{wang2023far,wang-etal-2023-self-instruct}, evaluation~\citep{suzgun-etal-2023-challenging,zheng2023judging}, hallucination~\citep{rawte2023troubling,zhang2023language}, bias~\citep{ladhak-etal-2023-pre,feng-etal-2023-pretraining}, detecting generated text~\citep{pmlr-v202-mitchell23a} watermarking~\citep{pmlr-v202-kirchenbauer23a}, poisoning~\citep{pmlr-v202-wan23b}, detecting pre-training data~\citep{shi2023detecting}, unlearning~\citep{yao2023large}, and efficient inference~\citep{dettmers2023qlora}.

While there are various motivations, such as raising the level of research and development in natural language processing, elucidating the mechanisms of LLM intelligence, the security risks of relying on a handful of foreign companies, and achieving responsible artificial intelligence, several Japanese companies and universities have actively been developing open LLMs that achieve good performance on Japanese text.
However, in many cases, the quality of the training data for building Japanese LLMs was not satisfactory because they were mostly developed overseas as a part of multilingual corpora.

\begin{table*}[p]
    \centering \small
    \begin{tabular}{p{9.9em}|rrp{9em}p{4.9em}p{4.5em}}
        \hline
        \multicolumn{1}{c|}{\textbf{Corpus}} & \multicolumn{1}{c}{\textbf{Size (en)}} & \multicolumn{1}{c}{\textbf{Size (ja)}} & \multicolumn{1}{c}{\textbf{Source}} & \multicolumn{1}{c}{\textbf{Text}} & \multicolumn{1}{c}{\textbf{Langdet}} \\
        \hline
        \verb|statmt.org/ngrams| & 976 BT & --- & 2012, 2013 & WET? & \verb|cld2| \\
        \citep{buck-etal-2014-n} & \multicolumn{5}{|p{30em}}{\textbf{Dedup}: Remove lines with exact hash values (MurmurHash, 64 bit).} \\
        & \multicolumn{5}{|p{30em}}{\textbf{Clean}: Remove email addresses and HTML fragments; normalize punctuation and capitalization (Moses truecaser).} \\
        \hline
        CC-100 (CCNet) & 532 BT & 26 BL & 2018 & WET & fastText \\
        \citep{wenzek-etal-2020-ccnet} & \multicolumn{5}{|p{30em}}{\textbf{Dedup}: Remove paragraphs with exact hash values (SHA-1, 64 bit).} \\
        & \multicolumn{5}{|p{30em}}{\textbf{Clean}: Use documents with high likelihood computed by the language model trained on Wikipedia.} \\
        \hline
        C4 & 156 BT & --- & April 2019 & WET & \verb|langdetect| \\
        \citep{JMLR:v21:20-074} & \multicolumn{5}{|p{30em}}{\textbf{Dedup}: Remove exact matches in three-sentence units.} \\
        & \multicolumn{5}{|p{30em}}{\textbf{Clean}: Keep only lines ending in a punctuation mark; remove documents with one or two sentences; remove lines with 4- words; etc.} \\
        \hline
        mC4 & 2,733 BT & 240 BL & 71 months & WET & \verb|cld3| \\
        \citep{xue-etal-2021-mt5} & \multicolumn{5}{|p{30em}}{\textbf{Dedup}: Same as C4} \\
        & \multicolumn{5}{|p{30em}}{\textbf{Clean}: Same as C4, but they use a modified rule for keeping text (because the punctuation filter of C4 is specific to English): the text must be at least 200 characters long and contain at least 3 lines.} \\
        \hline
        OSCAR 23.01 & 523 BW & 74 BL & Nov/Dec 2022 & WET & fastText \\
        \citep{abadji-etal-2022-towards} & \multicolumn{5}{|p{30em}}{\textbf{Dedup}: TLSH} \\
        & \multicolumn{5}{|p{30em}}{\textbf{Clean}: Remove documents with high likelihood computed by the language models trained on harmful content or with URLs registered in Blacklists UT1.} \\
        \hline
        Pile-CC & 2,312 BL & --- & 2013--2020 & \verb|jusText| & \verb|cld2| \\
        \citep{gao2020pile} & \multicolumn{5}{|p{30em}}{\textbf{Dedup}: MinHash (Jaccard coefficient threshold is approximately 0.5).} \\
        & \multicolumn{5}{|p{30em}}{\textbf{Clean}: Use documents similar to those in OpenWebText2 (documents with a Reddit score of 3 or higher) based on a fastText classifier.} \\
        \hline
        ROOTS & 484 BB & --- & (539 domains) & (original) & fastText \\
        \citep{laurencon2023bigscience} & \multicolumn{5}{|p{30em}}{\textbf{Dedup}:  SimHashLSH (6-gram)} \\
        & \multicolumn{5}{|p{30em}}{\textbf{Clean}: Remove documents with too few words, with a high percentage of repeated characters, with a high percentage of repeated words, with too many emojis, or with too few function words, etc.} \\
        \hline
        RefinedWeb & 600 BT & --- & Until Jun 2023 & \verb|trafilatura| & fastText \\
        \citep{penedo2023refinedweb} & \multicolumn{5}{|p{30em}}{\textbf{Dedup}: MinHash (character 5-gram, approximate threshold 0.8)}. \\
        & \multicolumn{5}{|p{30em}}{\textbf{Clean}: Blacklist URL filter by UT1; remove repetitions; per-document filter; per-line modification; remove lines matching more than 50 tokens based on suffix array.} \\
        \hline
        Dolma & 2,415 BT & --- & May 2020 to Jun 2023 & WET & fastText \\
        \citep{DolmaDataset} & \multicolumn{5}{|p{30em}}{\textbf{Dedup}: Remove documents with identical URLs; remove identical paragraphs within documents.} \\
        & \multicolumn{5}{|p{30em}}{\textbf{Clean}: Paragraph filtering of MassiveWeb~\citep{rae2021scaling}; punctuation rules of C4; etc.} \\
        \hline
        ClueWeb 22 & *16.7 TT & 3,301 BT & Search engine & \verb|BlingFire| & (original) \\
        \citep{ClueWeb22} (Non-commercial use) & \multicolumn{5}{|p{30em}}{\textbf{Dedup} and \textbf{Clean}: Documents were sampled from a commercial search engine based on the distribution of page visits.} \\
        \hline
        RedPajama-Data-v2 & 20.5 TT & --- & All (84 dumps) & WET & fastText \\
        & \multicolumn{5}{|p{30em}}{\textbf{Dedup}: Same as CCNet }\\
        & \multicolumn{5}{|p{30em}}{\textbf{Clean}: Text-quality estimation rules used by C4, RefinedWeb, etc.}\\
        \hline
        This study (raw) & 3,684 BL & 3,684 BL & 2020-40 to 2023-23 & \verb|trafilatura| & (original) \\
        & \multicolumn{5}{|p{30em}}{\textbf{Dedup} and \textbf{Clean}: None.}\\
        \hline
        This study (clean) & 312 BL & 312 BL & 2020-40 to 2023-23 & \verb|trafilatura| & (original) \\
        & \multicolumn{5}{|p{30em}}{\textbf{Dedup}: MinHash (character 5-gram, Jaccard coefficient threshold is approximately 0.9).}\\
        & \multicolumn{5}{|p{30em}}{\textbf{Clean}: Repetition detection of RefinedWeb and quality filter specially designed for Japanese text.}\\
        \hline
    \end{tabular}
    \caption{Representative corpora that can be used to pre-train LLMs. BL, BW, BT, and TT stand for billion letters, billion words, billion tokens, and trillion tokens, respectively. Extracted from the corresponding papers, numbers in English size are not directly comparable because of the difference in units. The number of ClueWeb22 may include text in other languages. Numbers in Japanese size are in Unicode characters and are directly comparable. \textbf{Text}, \textbf{Langdet}, \textbf{Dedup}, and \textbf{Clean} explain methods for text extraction, language detection, deduplication, and cleaning.
    WET in text extraction indicates that the corpus uses the text extraction results distributed by Common Crawl in WET format.}
    \label{tab:cc-datasets}
\end{table*}

Table \ref{tab:cc-datasets} lists representative corpora used for training LLMs.
Some popular corpora such as Pile-CC~\citep{gao2020pile}, ROOTS~\citep{laurencon2023bigscience}, Dolma~\citep{DolmaDataset}, and RedPajama-Data-v2\footnote{\url{https://www.together.ai/blog/redpajama-data-v2}} cannot be used for training Japanese LLMs, including no Japanese text.
Therefore, CC-100~\citep{wenzek-etal-2020-ccnet}, mC4~\citep{xue-etal-2021-mt5}, and OSCAR 23.01~\citep{abadji-etal-2022-towards} are candidates for training Japanese LLMs, including a certain amount of Japanese text and released with permissive licenses.
However, the quality of Japanese text in these corpora is unsatisfactory because they contain noise in the HTML-to-text conversion (Common Crawl WET format) and because they have incorporated no special efforts to improve text quality for Japanese.
Although we recognize the importance of efforts in building multilingual corpora, it is difficult to build a useful and reliable Japanese corpus without the knowledge of Japanese language.

Therefore, this paper explores a method to construct a large-scale, high-quality Japanese web corpus that can be used for training Japanese LLMs.
The presented method includes lightweight language detection that boosts the processing speed of text extraction in the target language. This strategy is applicable not only to Japanese but also to other languages with less text than English.
In addition, we specially design a filtering method to find good-quality Japanese text.
The corpus developed in this study is made from 21 snapshots of Common Crawl (from CC-MAIN-2020-40 to CC-MAIN-2023-23), and the size of the corpus after cleaning is 173,350,375 pages and 312,093,428,689 characters.
To confirm the quality of the corpus, we perform continual pre-training on Llama 2 7B, 13B, 70B~\citep{touvron2023llama}, Mistral 7B v0.1~\citep{jiang2023mistral}, and Mixtral 8x7B Instruct~\citep{jiang2024mixtral} as base LLMs. Experimental results demonstrate that continual pre-training consistently improves the base model's performance by 6.6--8.1 points on Japanese benchmark datasets.
We also demonstrate that the improvement on Llama 2 13B brought from the presented corpus was the largest among those from other existing corpora.
The models trained on the presented corpus are available on Hugging Face\footnote{\url{https://huggingface.co/tokyotech-llm}}.

\section{Related work}
\label{sec:related-work}

A number of large corpora were built from Common Crawl\footnote{\url{https://commoncrawl.org/}} archives.
Common Crawl is a non-profit organization that crawls websites and provides their archives.
The crawled data was initially distributed in ARC format\footnote{\url{https://archive.org/web/researcher/ArcFileFormat.php}}, but since the summer of 2013, it has been stored in Web ARChive (WARC)\footnote{\url{https://iipc.github.io/warc-specifications/specifications/warc-format/warc-1.1/}} and Web Text (WET) formats.
According to Statistics of Common Crawl Monthly Archives\footnote{\url{https://commoncrawl.github.io/cc-crawl-statistics/}}, the total amount of accessible archives (data crawled since 2013 till 2023) is 251,325,355,174 pages.
Compact Language Detector 2 (\verb|cld2|)\footnote{\url{https://github.com/CLD2Owners/cld2}} estimated that about 5 percent of these web pages are written in Japanese.

Table \ref{tab:cc-datasets} summarizes the corpora derived from Common Crawl (except for ClueWeb22, where documents are sampled from a commercial search engine).
All permissive corpora that contain Japanese text rely on the data in WET format (HTML pages were converted to text by Common Crawl).
The advantage of processing data in WET format is the reduction of processing time and data transfer size.
\citet{wenzek-etal-2020-ccnet} proposed a pipeline to extract multilingual text from Common Crawl to train a multilingual BERT model (XLM-R)~\citep{conneau-etal-2020-unsupervised}, and released its implementation (CCNet)\footnote{\url{https://github.com/facebookresearch/cc_net}} and data (CC-100)\footnote{\url{https://huggingface.co/datasets/cc100}}.
\citet{xue-etal-2021-mt5} constructed mC4, a multilingual extension of C4~\citep{JMLR:v21:20-074} to train mT5, a multilingual variant of T5.
\citet{abadji-etal-2022-towards} released OSCAR 23.01 with adult content filtering (based on an $n$-gram language model) and duplicate removal (based on locality sensitive hashing) added to the previous release.

However, the corpus construction procedure often introduces aggressive filtering rules because the WET data usually includes irrelevant text, e.g., noises in HTML-to-text conversion, JavaScript codes, navigation menus, and footers.
For example, C4~\citep{JMLR:v21:20-074} had to introduce filtering rules such as ``remove lines with the word \emph{JavaScript}'' (to remove JavaScript code) and ``remove pages with curly brackets \verb|{}|'' (because curly brackets are rarely used in a natural language but often used in a programming language).
Therefore, we use WARC instead of WAT as the source format of Common Crawl archives.

We follow the design principle of RefinedWeb~\citep{penedo2023refinedweb}: \emph{scale first} (avoid labour intensive human curation process), 
\emph{strict deduplication} (respect the value of deduplication for large language models), and \emph{neutral filtering} (avoid undesirable biases introduced by machine-learning-based filtering).
The construction procedure of RefinedWeb has much in common with MassiveWeb~\citep{rae2021scaling}.
Collecting web pages with their own web crawler, \citet{rae2021scaling} used Google's SafeSearch filter to remove harmful content, and extracted text based on the HTML DOM structure.
They also removed documents that contained a lot of repetitions, low-quality text, or duplicated information.
In this study, we extend their pipelines to extract high-quality Japanese documents efficiently.

We also considered using ClueWeb22~\citep{ClueWeb22} as a source for building a large Japanese corpus.
Although ClueWeb22 has nice and unique properties such as real distribution (web pages are extracted by the crawler of a commercial search engine and follow the distribution of web search activities), large-scale quality content (content extraction pipeline based on the search engine's production-quality content understanding system), and rich information (annotations of content structure, in-links/out-links of web pages, visual features of contents, etc), we could not use ClueWeb22 in this study because an LLM trained on ClueWeb22 cannot be released under an open license.

\section{Method}
\label{sec:method}

\begin{figure*}[!t]
    \begin{center}
      \includegraphics[width=0.9\textwidth]{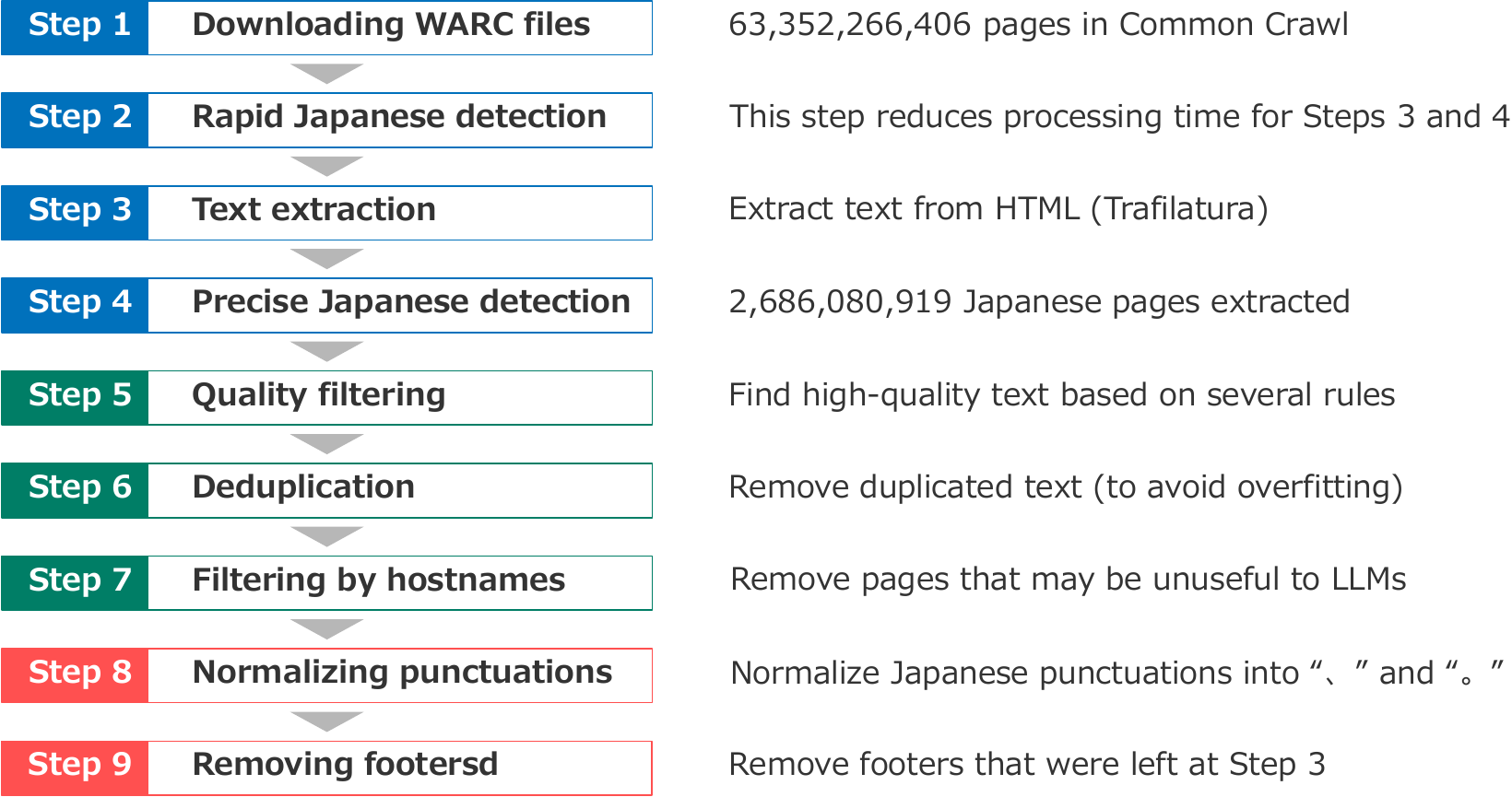}
    \end{center}
    \caption{The pipeline for building a large Japanese web corpus.}
    \label{fig:overview}
\end{figure*}

Figure \ref{fig:overview} shows the corpus building pipeline.
This pipeline is roughly divided into the following three stages: (1) Extracting Japanese text from WARC files in Common Crawl (Section \ref{sec:raw}); (2) Selecting Japanese text carefully with quality filtering (Section \ref{sec:quality-filtering}), deduplication (Section \ref{sec:dedup}), and host filtering (Section \ref{sec:host-filtering}); (3) Cleaning extracted text (Section \ref{sec:norm}). 

\subsection{Extracting Japanese text}
\label{sec:raw}

Common Crawl snapshots are stored as buckets in Amazon S3 and can be accessed via S3 or the web server\footnote{\url{https://data.commoncrawl.org/}}.
We extract HTML content from WARC data using the \verb|WARCIO| library\footnote{\url{https://github.com/webrecorder/warcio}} (Step 1, Figure \ref{fig:overview}).
We will defer to the explanation of Step 2 until later.
We then apply \texttt{Trafilatura}\footnote{\url{https://trafilatura.readthedocs.io/}}~\citep{barbaresi-2021-trafilatura} for extracting text from HTML markups (Step 3).
Step 4 detects the language of a text based on a linear binary classifier (see Appendix \ref{sec:langdet} for the detail) and extracts Japanese text only.

Because about 5\% of the entire Common Crawl is written in Japanese, it is possible to reduce the processing time of Steps 3 and 4 by 95\% if we can target Japanese web pages only.
However, in order to improve the accuracy of language detection, it is desirable to remove HTML markup before applying the language detection.
Therefore, we should apply text extraction first (Step 3), followed by language detection (Step 4).
To make matters worse, text extraction takes more processing time than language detection; we cannot reduce the processing time in this processing order.

This is the motivation behind Step 2: we select web pages that are highly likely to be in Japanese in a rapid manner without text extraction.
A web page proceeds to Steps 3 and 4 only when one of the following two conditions is met: (1) the HTML declares the language as Japanese, for example, \verb|<html lang="ja">|; (2) the precise Japanese detection (Section \ref{sec:langdet}) recognizes the content within \verb|<title>| tag as Japanese.

To evaluate the validity of this rule, we compared the results of rapid and precise language detection.
More specifically, assuming that the result of the precise language detection is correct, we measured the accuracy of rapid language detection on 10 WARC files\footnote{CC-MAIN-20230527223515-20230528013515-0000?.warc.gz from 0 to 9 in CC-MAIN-2023-23.} (394,192 pages, 12,052,992,707 bytes in total with gzip compression).
The precision, recall, and F1 score were 0.888, 0.967, and 0.926, respectively.
This result indicates that although rapid Japanese language detection may discard about 3.3\% of Japanese web pages, the relatively high precision reduces the processing time for Steps 3 and 4 for non-Japanese web pages.
In a benchmark experiment using 1 CPU of Intel Xeon Gold 6130 processor (2.1 GHz), the processing time for Steps 1--4 was about 15 times faster than Steps 1, 3, and 4 (without rapid language detection) because Step 2 removes a number of web pages from the target.

This stage obtained 2,686,080,919 pages and 3,684,320,219,695 Japanese characters from 21 snapshots of Common Crawl (from CC-MAIN-2020-40 to CC-MAIN-2023-23).

\subsection{Quality filtering}
\label{sec:quality-filtering}

This process (1) removes web pages with many repetitions, (2) selects web pages that contain good-quality Japanese text, and (3) removes web pages that may contain harmful expressions.
For processing (1), we adopted the rules of \citet{rae2021scaling} to remove documents with many duplicated contents (see Appendix \ref{sec:repetition-removal}).

As for processing (2), no standard has been established for assessing the quality of Japanese text.
Therefore, we designed several rules for (2) selecting Japanese text of good quality.
This study removes a text that satisfies either of the following conditions as irrelevant for LLMs (thresholds in parentheses).
\begin{enumerate}
\item Number of characters (less than 400 characters)
\item Percentage of hiragana characters (less than 0.2)
\item Percentage of katakana characters (greater than 0.5)
\item Percentage of Japanese characters (hiragana, katakana, kanji, punctuations) (less than 0.5)
\item Average number of characters in a sentence (less than 20 or more than 90)
\item Number of characters in the longest sentence (more than 200)
\item Percentage of sentences ending with an ellipsis symbol (greater than 0.2)
\end{enumerate}
For example, Rule 2 ensures that a text includes a certain amount of function words such as \emph{ga} (subject case marker), \emph{wo} (object case marker) \emph{ni} (position case maker roughly corresponding to \emph{to} in English).
Rule 3 rejects a web page containing a lot of product or service names (e.g., found in SEO sites), which are often expressed in katakana characters.
Rule 7 removes a web page that looks like an RSS feed (a collection of snippets of other web pages).
These rules were adjusted manually as the authors examined the text to be deleted and extracted.

Although RefinedWeb used UT1 blocklist\footnote{\url{https://dsi.ut-capitole.fr/blacklists/}} for removing harmful content, its coverage for Japanese pages may be insufficient.
In order to remove web pages that may contain harmful expressions (3), we manually created a list of NG expressions.
If the percentage of characters that match NG expressions is greater than 0.05, we exclude the text from the corpus.

This quality filtering reduced the size of the corpus to 646,238,066 pages (1,202,868,044,631 characters).
Examining the text before and after the quality filtering, we observe that web pages that may be unuseful for training LLMs (e.g., e-commerce sites) have disappeared.

\subsection{Deduplication}
\label{sec:dedup}

Because Common Crawl visits and crawls the same website multiple times, the archive includes web pages that are identical or similar because of minor modifications or reproductions.
\citet{lee-etal-2022-deduplicating} reported that deduplication, removing duplicated text from the corpus, not only reduces memorization of LLMs but also improves pre-training efficiency.

Therefore, we performed deduplication using MinHash~\citep{Broder:97} in a similar manner to \citet{lee-etal-2022-deduplicating}.
When using MinHash for deduplication, it is common to create $r$ buckets, where each bucket is a concatenation of $b$ MinHash values, compare $r$ pairs of buckets of two documents, recognize the two documents as a duplicate if any of bucket pairs are identical.
In this study, we adopted a setting where $b=20, r=40$ so that a pair of documents with a Jackard coefficient\footnote{In this study, features were constructed with $5$-grams of characters.} of $0.9$ can be approximately detected as a duplicate with a probability of $92.5$\%.
When a pair of documents was recognized as a duplicate, we kept the one crawled recently and removed the older one.
In the snapshot used in this experiment, the non-duplicate rate of web pages collected between March and June 2023 ranged from 77.8 to 87.9 percent, while the non-duplicate rate of web pages collected before February 2023 dropped to less than 40 percent, and the rate of web pages collected around 2020 was less than 20 percent.
This deduplication process reduced the corpus size to 185,766,463 pages (353,191,872,805 characters).

\subsection{Filtering by hostnames}
\label{sec:host-filtering}

Even after the quality filtering in Section \ref{sec:quality-filtering}, we found some irrelevant content in the corpus.
Therefore, we created a block list consisting of hostnames that met the following criteria.
\begin{enumerate}
    \item Included in the UT1 blocklist.
    \item Percentage of pages containing the name of a dating site (greater than 0.001).
    \item Percentage of pages containing NG expressions (greater than 0.005).
    \item \texttt{*wikipedia.org}
    \item \texttt{*.5ch.net}
\end{enumerate}

Although the UT1 blocklist includes some Japanese websites, we introduced Criteria 2 and 3 to improve the coverage.
The authors manually adjusted the thresholds for the percentage\footnote{We placed emphasis on recall at the expense of precision to reduce harmful behaviors of trained models. This design results in including a lot of non-harmful websites in the block list.}.
We applied Criterion 4 because we planned to use Wikipedia text extracted from the dump.
This filtering process resulted in a corpus of 173,350,375 pages (312,674,522,606 characters).

\subsection{Cleaning extracted text}
\label{sec:norm}

Sections \ref{sec:raw} to \ref{sec:host-filtering} process text at the document level, i.e., filter out irrelevant documents without altering the text within a document.
In this study, to avoid an unexpected side effect, we minimally edit the text: punctuation normalization (replacing Western-style punctuations with Japanese-style ones) and footer trimming (removing footers that were left by \texttt{Trafilatura}).
Refer to Appendix \ref{sec:norm-detail} for the details.
The process does not delete web pages, but the number of characters decreased from 	312,674,522,606 to 312,093,428,689.

\section{Experiments}
\label{sec:experiment}

\subsection{Models and training data}

In order to evaluate the usefulness of the presented corpus, we conducted continual pre-training of popular LLMs that excel in English\footnote{It may be straightforward to evaluate the corpus by training LLMs from scratch, but we chose continual pre-training because the broader goal of this effort was to build high-performance LLMs that excel in Japanese under the limited amount of computing budget. We can assess the quality of the corpus by checking performance gains from the base models and also comparing the performance with other models.}.
We used Llama 2 7B, 13B, 70B, Mistral 7B v0.1, and Mixtral 8x7B Instruct v0.1 as base models and performed continual pre-training of these base models on the presented corpus to examine whether the corpus improves their Japanese knowledge and skill.
The training data for continual pre-training was a mixture of the presented corpus, Japanese Wikipedia, RefinedWeb, and arXiv component in The Pile\footnote{We expected that the RefinedWeb and arXiv would help maintain the base LLMs' English knowledge and skill during continual pre-training.}~\citep{gao2020pile}.
Specifically, we prepared approximately 104.9 BT of training data, assuming a sequence length of 4,096 tokens and a batch size of 1,024 with 25,000 steps for continuous pre-training.
The ratio of Japanese to English tokens was set to 9:1, with 5\% of the training data being the English text from RefinedWeb, 5\% being The Pile's arXiv paper text (English), and the remaining 90\% being Japanese text.
The breakdown of the Japanese text was about 1.6 BT of Japanese Wikipedia, and the presented Japanese corpus occupied the rest.
We also used AlgebraicStack~\citep{azerbayev2024llemma-algebraicstack} for Mistral's continual pre-training. For Mixtral, we used both AlgebraicStack and The Vault~\citep{manh-etal-2023-vault}\footnote{We completed the continual pre-training experiments on Llama 2 well before we began with Mistral 7B. By then, our objective had expanded to enhance logical inference capabilities in Mistral and Mixtral, leading us to incorporate AlgebraicStack and The Vault.}.
We used a different ratio (72:28) of Japanese to English tokens for continual pre-training on Mixtral 8x7B Instruct\footnote{We changed this ratio as an attempt to improve the performance in English.}.

We did not change the architecture of base LLMs; the embedding sizes of tokens, hidden layers, feed-forward layers, the number of attention heads, and the number of layers were unchanged from the base models in continual pre-training.
Before continual pre-training, we added Japanese vocabulary to Llama 2 and Mistral 7B tokenizers.
The total number of vocabulary was 43,176 for LLMs based on Llama 2 and 42,800 for those based on Mistral 7B.
We employed AdamW~\citep{loshchilov2019decoupled}, with hyperparameters set to $\beta_1=0.9$, $\beta_2=0.95$, $\epsilon=1.0 \times 10^{-8}$.
We used re-warming and re-decaying the learning rate~\citep{ibrahim2024simple-rewarming} with 1,000 warm-up steps. For Llama 2, the maximum learning rate was set at $1.0 \times 10^{-4}$ with a decay rate of $1/30$, while for Mistral and Mixtral, it was $2.0 \times 10^{-5}$ with a decay rate of $1/10$. The batch size was 1,024.
We used $0.1$ for weight decay and 1.0 for gradient clipping.
In addition, we used Flash Attention~\citep{NEURIPS2022_67d57c32} to improve computational and memory efficiency.

In order to examine the impact of a corpus on the performance of LLMs, we performed continual pre-training on ClueWeb22 and llm-jp-corpus v1.0.1\footnote{\url{https://github.com/llm-jp/llm-jp- corpus}} as well as the presented corpus.
We also compared our LLMs to others with similar numbers of parameters: CyberAgentLM2 7B (CALM2 7B, a Japanese LLM trained from scratch); Japanese-StableLM-Base-Beta-7B (JSLMB 7B, a continual pre-training LLM on Llama 2 7B); Youri 7B (a continual pre-training LLM on Llama 2 7B); Qwen 7B; Nekomata 7B (a continual pre-training LLM on Qwen 7B); Japanese Stable LM Base Gamma 7B (JSLMG 7B, a continual pre-training LLM on Mistral 7B); Qwen 14B; KARAKURI LM 70B (Karakuri 70B, a continual pre-training LLM on Llama 2 70B); Japanese-StableLM-Base-Beta-70B (JSLMB 70B, a continual pre-training LLM on LLama 2 70B); and Qwen 72B.
Refer to Appendix \ref{sec:llms} for the complete list of URLs of the LLMs used in the experiments.

\subsection{Evaluation Dataset}
We used llm-jp-eval\footnote{\url{https://github.com/llm-jp/llm-jp-eval} v1.0.0} and lm-evaluation-harness\footnote{\url{https://github.com/Stability-AI/lm-evaluation-harness/tree/jp-stable} commit \#9b42d41} as Japanese evaluation benchmarks.
llm-jp-eval is a benchmark consisting of tasks for multiple choice (MC) question answering, open question answering (QA), reading comprehension (RC), and natural language inference (NLI) tasks.
JCommonsenseQA~\citep{kurihara-etal-2022-jglue} is employed for MC, JEMHopQA~\citep{ishi-etal-2023-jemhopqa} and NIILC~\citep{sekine-etal-2003-niilc} for QA, JSQuAD~\citep{kurihara-etal-2022-jglue} for RC.
We decided to exclude the NLI dataset from the benchmark because the distribution of ground-truth NLI labels is highly imbalanced, which made the evaluation unstable\footnote{An LLM may obtain a high score only if the model happens to predict the majority label without understanding and solving the NLI task.}.
For lm-evaluation-harness, we used Japanese subsets of XL-Sum~\citep{hasan-etal-2021-xlsum} for the automatic summarization task and MGSM~\citep{shi-etal-2022-mgsm} for the arithmetic reasoning task, respectively.
In addition, WMT 2020~\citep{barrault-etal-2020-findings-wmt20} was used for the Japanese-English and English-Japanese machine translation.

\subsection{Results}

\begin{table*}[t]
    \centering \small
    \begin{tabular}{l|rrrrrrrr|r}
    \hline
    \multicolumn{1}{c}{Corpus} & \multicolumn{1}{|c}{QA} & \multicolumn{1}{c}{QA} & \multicolumn{1}{c}{QA} & \multicolumn{1}{c}{RC} & \multicolumn{1}{c}{Sum} & \multicolumn{1}{c}{Ja-En} & \multicolumn{1}{c}{En-Ja} & \multicolumn{1}{c|}{Math} & \multicolumn{1}{c}{Avg} \\
    & \multicolumn{1}{c}{\scriptsize JCom} & \multicolumn{1}{c}{\scriptsize JEMHop} & \multicolumn{1}{c}{\scriptsize NIILC} & \multicolumn{1}{c}{\scriptsize JSQuAD} & \multicolumn{1}{c}{\scriptsize XL-Sum} & \multicolumn{1}{c}{\scriptsize WMT20} & \multicolumn{1}{c}{\scriptsize WMT20} & \multicolumn{1}{c|}{\scriptsize MGSM} & \\
    \hline
    CALM2 7B & 21.98 & 50.47 & 50.66 & 77.99 & 2.33 & 14.99 & 23.45 & 6.00 & 30.98 \\
    \hdashline
    JSLMB 7B & 36.10 & 44.78 & 44.32 & 83.18 & 21.95 & 12.26 & 19.46 & 7.20 & 33.66 \\
    Youri 7B & 46.20 & 47.76 & 49.99 & 85.06 & 19.57 & \textbf{19.71} & 26.71 & 6.40 & 37.67 \\    
    Llama 2 7B & 38.52 & 42.40 & 34.10 & 79.17 & 19.05 & 17.37 & 17.83 & 7.60 & 32.01 \\
    + \emph{this study} & 48.08 & \textbf{50.78} & \textbf{59.68} & 85.73 & 18.30 & 15.11 & 25.10 & 12.40 & 39.40 \\
    \hdashline
    Qwen 7B & 77.12 & 42.34 & 23.76 & 85.94 & 13.71 & 18.01 & 16.89 & 21.60 & 37.42 \\
    Nekomata 7B & 74.17 & 49.28 & 50.22 & 87.07 & 16.76 & 18.15 & \textbf{26.73} & 12.40 & 41.85 \\
    \hdashline
    JSLMG 7B & 73.64 & 46.43 & 55.68 & \textbf{89.10} & \textbf{22.93} & 15.61 & 23.90 & 16.80 & 43.01 \\
    Mistral 7B & 73.01 & 42.45 & 27.22 & 85.63 & 20.06 & 17.33 & 14.05 & 17.60 & 37.17 \\
    + \emph{this study} & \textbf{85.70} & 49.15 & 55.19 & 88.02 & 19.88 & 16.67 & 24.94 & \textbf{22.40} & \textbf{45.24} \\
    \hline
    Llama 2 13B & 68.19 & 44.55 & 41.74 & 85.51 & 21.33 & 19.81 & 21.36 & 13.20 & 39.46 \\
    + ClueWeb22 & 76.76 & 49.29 & 56.02 & 89.55 & 20.15 & \textbf{23.32} & \textbf{29.08} & 11.60 & 44.47 \\
    + llm-jp & 74.71 & \textbf{50.85} & 60.34 & \textbf{90.28} & \textbf{21.91} & 17.89 & 25.89 & 18.00 & 44.98 \\
    + \emph{this study} & \textbf{78.37} & 50.63 & \textbf{64.06} & 90.07 & 21.68 & 17.71 & 27.37 & \textbf{21.60} & \textbf{46.44} \\
    \hline
    Qwen 14B & 88.29 & 42.43 & 32.20 & 89.80 & 18.51 & 22.24 & 22.23 & 38.80 & 44.31 \\
    \hline
    Mixtral 8x7B & 84.00 & 50.33 & 31.07 & 88.08 & 20.02 & 20.63 & 19.56 & \textbf{45.20} & 44.86 \\
    + \emph{this study} & \textbf{92.58} & \textbf{58.43} & \textbf{56.87} & \textbf{91.48} & \textbf{25.89} & \textbf{20.74} & \textbf{27.05} & 43.60 & \textbf{52.08} \\
    \hline
    Karakuri 70B & 85.79 & 51.25 & 57.13 & 91.00 & 14.64 & 21.13 & 25.40 & 27.20 & 46.69 \\
    JSLMB 70B & 91.15 & 49.25 & 60.42 & \textbf{91.92} & \textbf{25.73} & 23.35 & 27.65 & 41.60 & 51.38 \\
    Llama 2 70B & 86.86 & 46.56 & 52.56 & 90.80 & 23.61 & \textbf{23.98} & 26.43 & 35.60 & 48.30 \\
    + \emph{this study} & \textbf{93.48} & \textbf{62.90} & \textbf{69.60} & 91.76 & 22.66 & 22.98 & \textbf{30.43} & \textbf{48.40} & \textbf{55.28} \\
    \hline    
    Qwen 72B & 92.94 & 55.66 & 45.18 & 91.59 & 21.79 & 23.56 & 25.61 & 63.20 & 52.44 \\
    \hline
    \end{tabular}
    \caption{Benchmark evaluation on Japanese tasks. A horizontal line groups LLMs with the same number of parameters. A horizontal dash line groups LLMs from the same base model. A bold number indicates the maximum value in the LLMs with the same number of parameters.}
    \label{tab:japanese-evaluation}
\end{table*}

Table \ref{tab:japanese-evaluation} reports the performance of LLMs on the Japanese benchmark datasets. The row ``+ \emph{this study}'' shows the performance of the LLM after we applied continual pre-training to the base model on the presented corpus.
The rows ``+ \emph{this study}'' demonstrate that the presented corpus consistently improves the performance of the base models via continual pre-training by 6.6--8.1 points on average. The models built by this study outperform other models built from scratch or continual pre-training, establishing the state-of-the-art performance in each model size (7B, 13B, 8x7B, and 70B).
This fact shows the usefulness of the presented corpus in building LLMs that excel in Japanese.

We can compare different Japanese corpora in continual pre-training of Llama 2 13B.
All corpora (ClueWeb22, llm-jp-corpus v1.0.1, and presented corpus) improved the base model's performance by 5.0--7.0 points, which verifies the effectiveness of the continual pre-training.
The presented corpus yielded the best improvement (7.0 points) of all corpora; in particular, it drastically enhances the performance of question answering (JCom, JEMHop, NIILC), reading comprehension (JSQuAD), and arithmetic reasoning (MGSM).
Although we did not incorporate a special effort to improve these tasks, the model acquired knowledge about Japan and the Japanese language from the presented corpus.
In contrast, the presented corpus did not improve the summarization (XL-Sum) task. This trend is observed across other base models, except for Mixtral 8x7B Instruct, potentially indicating that adding Japanese vocabulary may have a detrimental effect on this task.
In addition, we found that the continual pre-training on the presented corpus improved the performance in English-Japanese translation but degraded that in the reversed direction (Japanese-English translation).
We will explore mitigation of this phenomenon in the future, e.g., by changing the ratio of Japanese to English tokens in the training data and promoting language transfer using an English-Japanese parallel corpus during continual pre-training.

\section{Conclusion}
\label{sec:conclusion}

In this paper, we built a large Japanese web corpus by extracting and refining text from the Common Crawl archive (21 snapshots of approximately 63.4 billion pages crawled between 2020 and 2023).
This corpus consists of approximately 312.1 billion characters (approximately 173 million pages), which is the largest of all available training corpora for Japanese LLMs.
We confirmed the usefulness of the corpus by performing continual pre-training on Llama 2 7B, 13B, 70B, Mistral 7B v0.1, and Mixtral 8x7B Instruct as base LLMs.
The experiments demonstrated consistent (6.6--8.1 points) improvements in Japanese benchmark datasets, and established the state-of-the-art performance in each model size (7B, 13B, 8x7B, and 70B).
We also observed that the improvement on Llama 2 13B brought from the presented corpus was the largest among those from other existing corpora.

Future directions include efforts towards the safety of LLMs, such as reducing harmful generations (e.g., discrimination, exclusion, toxicity, hallucination).
Currently, we only use the lists of NG expressions and hostnames, but it is desirable to establish more robust filtering methods to remove harmful text for pre-training Japanese LLMs.
In addition, although our study focused on the continual pre-training setting, we want to evaluate the presented corpus by training Japanese LLMs from scratch.
Although we evaluated the LLMs in downstream tasks such as question-answering and summarization, it is questionable whether this can measure the ``general intelligence'' of an LLM.
At the same time, training an LLM on a pre-training corpus requires huge computations.
Therefore, we want to explore a lightweight method for assessing the effectiveness of pre-training corpora without building LLMs.

\section{Limitations}
\label{sec:limitation}

We could not present an ablation study of each step in Figure \ref{fig:overview} because of a large amount of computational resources required to train LLMs.
Instead, we reported the performance of each step of the corpus construction procedure in Section \ref{sec:method}.

This study focuses on Japanese, and the construction of corpora for other languages is outside the scope. For English, it may be a good strategy to make use of good quality corpora that have already been developed. Although some ideas in this paper could be helpful in building corpora for other languages, for example, reducing processing time with rapid language detection, we believe that it would be better for researchers from different countries to share the task of building corpora for their own, as assessing text quality requires the knowledge of the language.

\section{Ethics statement}

Article 30-4 of the Copyright Act in Japan permits to use of a copyrighted work without the permission of the copyright holder as long as the use is not a person's purpose to personally ``enjoy'' or cause another person to ``enjoy'' the work.
This provides the justification for building the presented corpus and LLMs.
To the best of our knowledge, we have taken every measure to keep the corpus as non-toxic and unbiased as possible, but we are unaware of any direct ethical consequences caused by the LLMs trained on the presented corpus.

\section{Reproducibility statement}

All models developed in this study (continual pre-training on Llama 2 7B, 13B, 70B, Mistral 7B v0.1, and Mixtral 8x7B Instruct) have already been released on Hugging Face. The benchmark datasets used in this study are also publicly available. Therefore, it is straightforward to reproduce our experimental results reported in Table \ref{tab:japanese-evaluation}.

\section*{Acknowledgements}

This paper is based on results obtained from a project, JPNP18002, commissioned by the New Energy and Industrial Technology Development Organization (NEDO).
In addition, the experiments of continual pre-training of LLMs was supported by the ``Support Program for Building Large Language Models'' of the AI Bridging Cloud Infrastructure (ABCI) developed and operated by the National Institute of Advanced Industrial Science and Technology (AIST).
We used the datasets and findings released by the Japanese LLM Study Group (LLM-jp) in the evaluation experiments.
We also received suggestions on tokenization from Tatsuya Hiraoka of Fujitsu Ltd.

\bibliography{colm2024_conference}
\bibliographystyle{colm2024_conference}

\appendix
\section{Appendix}

\subsection{Precise language detection}
\label{sec:langdet}

Following the pipeline of RefinedWeb~\citep{penedo2023refinedweb}, we built a language detector with a linear discriminator using character $n$-grams as features.
We used multilingual Wikipedia texts as the training data for the linear discriminator, and the feature space was constructed from the training data with character unigrams, bi-grams, and trigrams that satisfy one of the following criteria.
\begin{enumerate}
\item Within the top 400,000 occurrences in the training data for all languages.
\item Within the top 400,000 occurrences in the training data of Japanese.
\item Within the top 100,000 occurrences in Chinese training data.
\item Within the top 10,000 occurrences in the training data for each language.
\end{enumerate}
If only Criterion (1) was used, there would be a possibility that features related to Japanese characters would not be obtained sufficiently. Therefore, Criterion (2) aimed to obtain features specific to Japanese. Criterion (3) made it easier to distinguish between Japanese and Chinese, which share Chinese characters. We aimed to stabilize the detection results for texts in languages with insufficient training data using Criterion (4).
The total number of distinct features was 821,484.

For training the linear discriminator, we used the dump of Wikipedia CirrusSearch\footnote{\url{https://dumps.wikimedia.org/other/cirrussearch/}} (as of June 12, 2023).
In training the language detector, we reduced the amount of training data to 1/20th of the dump because the total amount was too large for training the discriminator. We sampled 100,000 instances in the remaining data for development and evaluation sets.
We trained an L2-regularized L2-loss Support Vector Machine on the training data using \verb|LIBLINEAR|\footnote{\url{https://www.csie.ntu.edu.tw/~cjlin/liblinear/}} 2.46. The regularization coefficient was set to $C=10$ empirically by the search on the development set.
We confirmed that the performance of this classifier was quite good: 0.996 F1 score on the test set and 0.989 F1 score on \texttt{whatlang-corpora}\footnote{\url{https://github.com/whatlang/whatlang-corpora}}.

\subsection{Rules for removing repetitions}
\label{sec:repetition-removal}

The removal of documents that contain many repetitions aims to filter out documents that do not seem to have useful information and to prevent the behavior of LLMs from repeatedly generating the same words.
Specifically, we follow the rules of \citet{rae2021scaling} and remove a document when the percentage of any of the following exceeds the threshold (shown in parentheses).
\begin{enumerate}
\item Number of lines duplicated in other lines / total number of all lines (0.30)
\item Number of paragraphs that are duplicates of other paragraphs / total number of paragraphs (0.30)
\item Number of characters that appear in other lines / total number of all characters (0.20)
\item Number of characters in paragraphs that appear in other paragraphs / total number of characters (0.20)
\item Number of occurrences of the most frequent 2-gram / total number of occurrences of all 2-grams (0.20)
\item Number of occurrences of the most frequent 3-gram / total number of occurrences of all 3-grams (0.18)
\item Number of occurrences of the most frequent 4-gram / total number of occurrences of all 4-grams (0.16)
\item Total number of occurrences of 5-grams appearing more than once / Total number of occurrences of all 5-grams (0.15)
\item Total number of occurrences of 6-grams appearing more than once / Total number of occurrences of all 6-grams (0.14)
\item Total number of occurrences of 7-grams appearing more than once / Total number of occurrences of all 7-grams (0.13)
\item Total number of occurrences of 8-grams appearing more than once / Total number of occurrences of all 8-grams (0.12)
\item Total number of occurrences of 9-grams appearing more than once / Total number of occurrences of all 9-grams (0.11)
\item Total number of occurrences of 10-grams appearing more than once / Total number of occurrences of all 10-grams (0.10)
\end{enumerate}

\subsection{Clearning the text}
\label{sec:norm-detail}

In order to normalize the punctuation in the corpus to ``、'' and ``。'', we replaced ``，'' with ``、'' and ``．'' with ``。'', respectively, based on the following rule.
\begin{enumerate}
    \item Check the number of occurrences the symbols ``、'' and ``，'' appear in the document. If ``，'' appears more often than ``、'', replace ``，'' with ``、''. However, ``，'' preceding an alphanumeric character is excluded from the replacement with ``、''.
    \item  Check the number of occurrences the symbols ``。'' and ``．'' appear in the document. If ``．'' appears more often than ``。'', replace ``．'' with ``。''. However, ``。'' preceding an alphanumeric character is excluded from the replacement with ``．''.
\end{enumerate}
This punctuation normalization process replaced ``．'' with ``。'' in 290,318 documents (0.17\%) and ``，'' with ``、'' in 1,107,319 documents (0.64\%).

Although \texttt{Trafilatura} used for text extraction removes the navigation and footer text of web pages, we sometimes see footer text that have not be removed by \texttt{Trafilatura}.
Therefore, expressions such as "Trackback list for this article", "All rights reserved", and "Click" in the last three lines of text were removed from the text if they occupy more than 30\% of the text in character.
This footer removal process removed footers at the end of 12,617,787 documents (7.3\%) of the pages.

\subsection{List of LLMs used in the experiments}
\label{sec:llms}

\paragraph{Base models}

\begin{itemize}
    \item Llama 2 7B: \url{https://huggingface.co/meta-llama/Llama-2-7b-hf}
    \item Llama 2 13B: \url{https://huggingface.co/meta-llama/Llama-2-13b-hf}
    \item Llama 2 70B: \url{https://huggingface.co/meta-llama/Llama-2-70b-hf}
    \item Mistral 7B v0.1: \url{https://huggingface.co/mistralai/Mistral-7B-v0.1}
    \item Mixtral 8x7B Instruct v0.1:\\
    \url{https://huggingface.co/mistralai/Mixtral-8x7B-Instruct-v0.1}
\end{itemize}

\paragraph{Other models for comparison}

\begin{itemize}
\item CyberAgentLM2 7B: \url{https://huggingface.co/cyberagent/calm2-7b}
\item Japanese-StableLM-Base-Beta-7B:\\ \url{https://huggingface.co/stabilityai/japanese-stablelm-base-beta-7b}
\item Rinna Youri 7B: \url{https://huggingface.co/rinna/youri-7b}
\item Qwen 7B: \url{https://huggingface.co/Qwen/Qwen-7B}
\item Rinna Nekomata 7B: \url{https://huggingface.co/rinna/nekomata-7b}
\item Japanese Stable LM Base Gamma 7B:\\ \url{https://huggingface.co/stabilityai/japanese-stablelm-base-gamma-7b}
\item Qwen 14B: \url{https://huggingface.co/Qwen/Qwen-14B}
\item KARAKURI LM 70B: \\
\url{https://huggingface.co/karakuri-ai/karakuri-lm-70b-v0.1}
\item Japanese-StableLM-Base-Beta-70B:\\ \url{https://huggingface.co/stabilityai/japanese-stablelm-base-beta-70b}
\item Qwen 72B: \url{https://huggingface.co/Qwen/Qwen-72B}
\end{itemize}

\end{document}